\documentclass[11pt]{article}

\usepackage[preprint]{acl}

\usepackage{times}
\usepackage{latexsym}

\usepackage[T1]{fontenc}
\usepackage[utf8]{inputenc}

\usepackage{microtype}

\usepackage{inconsolata}

\usepackage{graphicx}

\usepackage{amsmath}
\usepackage{amssymb}
\usepackage{booktabs}
\usepackage{natbib}
\usepackage{amsmath}      % \text{} 용
\usepackage{algorithm}    % algorithm 환경 (float, caption)
\usepackage{algorithmic}  % \STATE, \FOR, \ENDFOR, \COMMENT
\usepackage{multirow}
\usepackage[table]{xcolor}  % \rowcolor
\usepackage{pgfplots}
\pgfplotsset{compat=1.18}

\title{Personality Latents: Conditioning Frozen Language Models on Individuals from Their Own Writing}

\newcommand\correspondingauthor{\thanks{Corresponding author.}}

\author{
 \textbf{Gunhee Cho\textsuperscript{1}},
 \textbf{Yun-Gyung Cheong\correspondingauthor\textsuperscript{1}}
\\
 \textsuperscript{1} Dept. of AI, Sungkyunkwan University
\\
  \textsuperscript{1} \{skate4333, aimecca\}@skku.edu
}

\begin{document}
\maketitle
\begin{abstract}
Controlling the personality of a language model usually relies on written profiles or
named trait categories, so it can only express what an author writes down or what a
taxonomy names. We instead propose \textbf{Personality Latents (PL)}, which learn a latent for
each person from their own writing, with no personality labels and no authored profile,
and inject it into a frozen language model as a soft prefix. Following generative
latent optimization, each training person receives a free latent that is optimized
through the frozen model itself, while an encoder learns to predict these latents so
that a new person needs only one forward pass, with optional refinement at test time.
We evaluate on $360$ held-out people by having the conditioned model answer neutral
and interview questions and reading personality from its answers with scorers trained
on real human writing. On the MBTI axes PL reaches an AUC of $0.782$, ahead of persona
embedding and prompt-based baselines under identical settings, and the ranking holds
when the answers are scored in three different embedding spaces. A latent learned with
Qwen2.5-3B-Instruct also transfers personality to Llama-3.1-8B-Instruct through a new
adapter alone.
\end{abstract}

\section{Introduction}
\label{sec:intro}

Controlling the personality of a language model has become a practical requirement.
Role playing agents must stay in character, assistants are expected to adapt to the
individual they serve, and simulated respondents are used to pilot studies before
human data is collected. Each of these needs a stable disposition, and personality is
the natural way to specify it because psychology has made it measurable.

Two representative approaches give a model personality. Profile based methods describe
a character in text and prompt or fine tune the model on that description
\citep{shao-etal-2023-character,wang-etal-2024-rolellm}, so what can be conveyed is limited to
what an author wrote and every new character needs its own training run. Trait based
methods condition on categories, through dialogue generated under Big Five levels
\citep{big5chat}, one model per MBTI type \citep{cui2023machine}, or steering
along a trait direction \citep{bhandari-etal-2026-activation}. Their resolution is that of the
taxonomy, so two people high on extraversion are treated identically and anything
the taxonomy does not name cannot be requested.

Persona embedding methods escape both limits by learning a vector from a person's own
behavior and injecting it into a frozen model
\citep{10.1145/3701716.3715463,liu-etal-2025-llms,qiu-etal-2025-latent}, or by moving this conditioning into per
user adapters \citep{tan-etal-2024-democratizing,tan-etal-2026-instant}. Three properties still stand between them and
personality. The person vector is constrained to be the output of an encoder, so the
search for a representation is confined to what that encoder can express. The vector
is learned together with one target model, which makes it a coordinate in that
model's input space rather than a portable description of the person. And these
methods are built for preference prediction and task accuracy on benchmarks such as
LaMP, with their authors noting that the embedding captures behavior rather than the
way a person writes \citep{liu-etal-2025-llms} and that evaluation has so far been confined
to a single setting \citep{qiu-etal-2025-latent}, so whether such a vector carries personality
remains open.

% We learn a personality latent from a person's own writing, with no labels and no
% authored profile, and inject it into a frozen backbone LLM. Following generative latent
% optimization \citep{glo}, each training person receives a free
% latent optimized jointly with the adapter through the backbone LLM, so the representation
% is not confined to what an encoder can produce. An encoder trained alongside predicts
% these latents for unseen people in one forward pass, and the same objective can refine
% a latent at test time. Because the objective does not tie the latent to one backbone LLM, a
% single vector can drive several models through their own small adapters. We evaluate
% by having the conditioned model write and then reading personality from that writing
% with scorers fitted on real human posts, comparing against the documented personality
% of people never seen in training.

We propose \textbf{Personality Latents (PL)}, which learn a personality latent from a person's
own writing, with no labels and no authored profile, and inject it into a frozen
backbone LLM. Following generative latent
optimization \citep{glo}, each training person receives a free
latent optimized jointly with the adapter through the backbone LLM, so the representation
is not confined to what an encoder can produce. An encoder trained alongside predicts
these latents for unseen people in one forward pass, and the same objective can refine
a latent at test time. Because the objective does not tie the latent to one backbone LLM, a
single vector can drive several models through their own small adapters. We evaluate
by having the conditioned model write and then reading personality from that writing
with scorers fitted on real human posts, comparing against the documented personality
of people never seen in training.

Our contributions are the following.

\begin{itemize}
\item \textbf{Individual level personality conditioning without labels or profiles.}
A latent estimated from a person's writing conditions a frozen backbone LLM on that person
rather than a category, with an amortizing encoder for unseen people and optional test
time refinement.
\item \textbf{A latent that is not tied to one backbone LLM.} The same frozen vector
transfers personality through a model never involved in learning it.
\item \textbf{Evaluation against documented personality.} We read personality from
what the conditioned model writes using scorers fitted on real human posts, compare
against documented MBTI and Big Five labels for held out people, and benchmark persona
embedding and prompt based methods under identical settings.
\end{itemize}

\section{Related Work}
\label{sec:related}

\subsection{Giving a persona to a language model}

Profile based methods write the persona down. Character-LLM converts a character's
profile into experiences and trains an agent on them
\citep{shao-etal-2023-character}, and RoleLLM builds role profiles for a hundred
characters and tunes models to speak in their style \citep{wang-etal-2024-rolellm}.
Because the description is authored, only what the author thought to record can be
conveyed, and the tuned variants need a separate training run for each character.

Trait based methods condition on categories drawn from a psychological taxonomy.
Big5-Chat generates dialogue under Big Five trait levels and trains on it
\citep{big5chat}, while Machine Mindset fine tunes one model per MBTI type through
supervised tuning followed by preference optimization \citep{cui2023machine}.
A related line edits the model's internal state instead of its weights, building a
direction in activation space from contrastive prompts and adding it during decoding
to raise or lower a named trait \citep{bhandari-etal-2026-activation}. All of these
take a label as the unit of control, so the conditioning is as fine grained as the
taxonomy and no finer. Our target is a particular individual rather than a category,
and no trait is named at any point.

\subsection{Persona embeddings}

Persona embedding methods represent the person as a vector learned from their own
behavior and inject it into a frozen model. USER-LLM pretrains a user encoder on
interaction histories and contextualizes the model through cross attention
\citep{10.1145/3701716.3715463}. Persona-Plug aggregates a user's history with weights
that depend on the current input, so the user vector changes from query to query
\citep{liu-etal-2025-llms}. DEP codes a user relative to peers who engaged with
similar content and compresses the result with a sparse autoencoder before injection
\citep{qiu-etal-2025-latent}. Others move the conditioning into parameters, training
an adapter per user \citep{tan-etal-2024-democratizing} or generating adapter weights
from a profile with a hypernetwork \citep{tan-etal-2026-instant}. Unlike these
methods, our latent is a free parameter optimized through the LLM rather than an
encoder output, is shared across LLMs through separate adapters, and is evaluated on
personality rather than on preference prediction.
% Method section.
% Requires: amsmath, algorithm/algorithmic (or algorithm2e), booktabs.

\begin{figure*}[t]
    \centering
    \includegraphics[width=\textwidth]{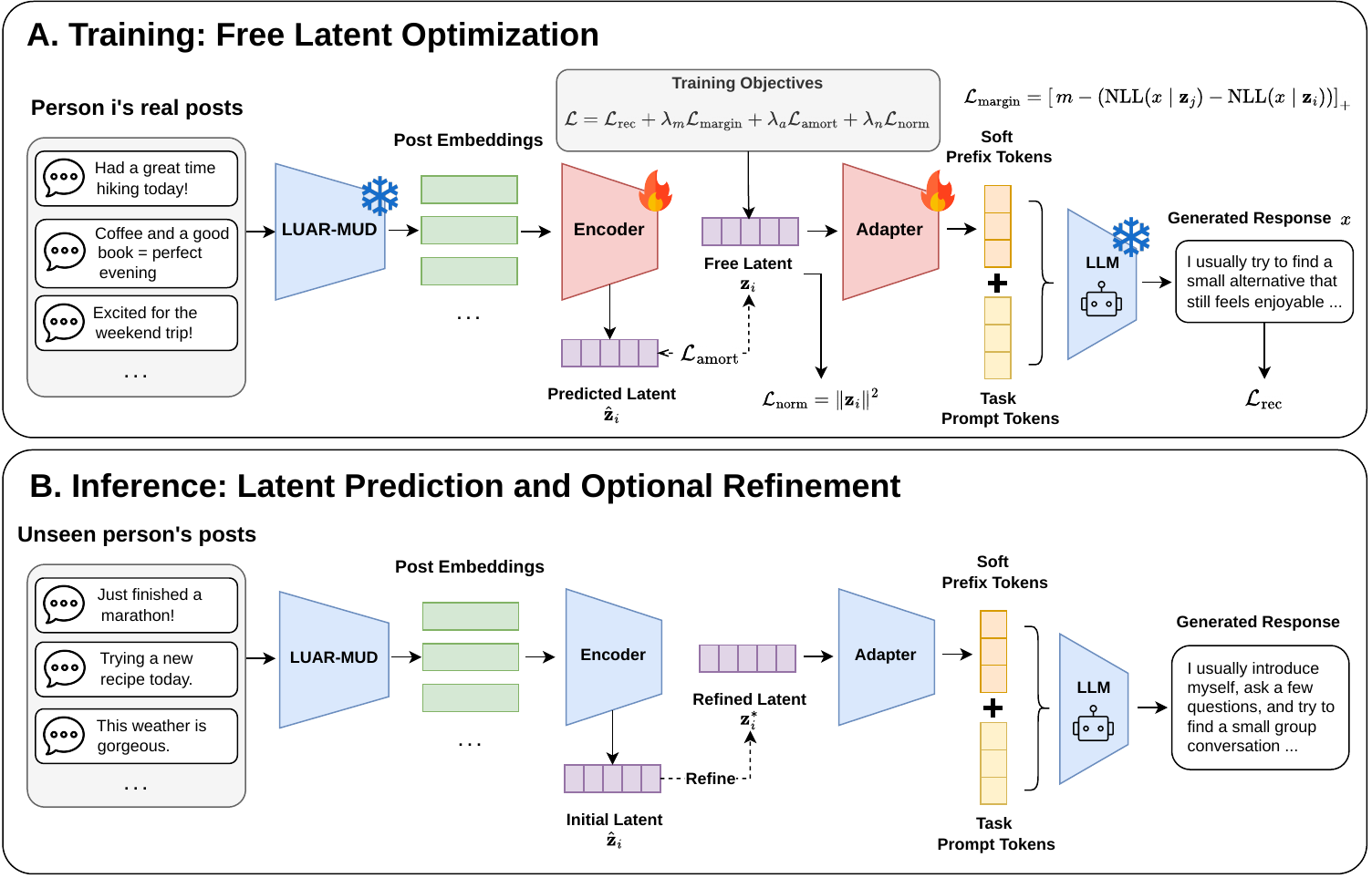}
    \caption{Overview of the proposed framework.
During training, a free person-specific latent is optimized through a frozen LLM,
while an amortizing encoder learns to predict it from the person's posts.
The latent is converted into soft prefix tokens by an adapter.
For an unseen person, the encoder prediction is used directly or optionally refined
at test time before generation.}
    \label{fig:persona-embedding}
\end{figure*}
\
\section{Method}
\label{sec:method}

We want a single low-dimensional vector that carries a person's personality and,
when injected into a \emph{frozen} language model, makes that model answer as that
person would. Let $\mathcal{P}$ denote a set of people, and let
$X_i = \{x_{i1},\dots,x_{iK_i}\}$ be the posts written by person $i$. We seek a
latent $\mathbf{z}_i \in \mathbb{R}^{d}$ and a light conditioning module such that
the frozen LLM, conditioned on $\mathbf{z}_i$, reproduces person $i$'s
personality. We call $\mathbf{z}_i$ the personality latent of person $i$, and refer to the
resulting method as \textbf{Personality Latents (PL)}. No personality labels are used at any point during training; the only
supervision is the person's own text.

\subsection{Conditioning a Frozen LLM}
\label{sec:adapter}

The latent is injected as a soft prefix. An adapter
$g_\theta: \mathbb{R}^{d} \rightarrow \mathbb{R}^{P \times d_{\text{model}}}$ maps
$\mathbf{z}$ to $P$ vectors in the LLM's input-embedding space,
\begin{equation}
g_\theta(\mathbf{z}) = \mathrm{reshape}\big(W_2\,\sigma(\mathrm{LN}(W_1 \mathbf{z}))\big),
\end{equation}
with $\sigma$ a GELU nonlinearity and $\mathrm{LN}$ layer normalization. These $P$
vectors are concatenated in front of the token embeddings of the prompt, and the
LLM runs on the concatenated sequence:
\begin{equation}
\mathbf{H} = \big[\, g_\theta(\mathbf{z})\;;\; E(\mathbf{t}) \,\big],
\qquad
p(y \mid \mathbf{t}, \mathbf{z}) = \mathrm{LM}(\mathbf{H}),
\end{equation}
% where $E$ is the decoder's (frozen) embedding table and $\mathbf{t}$ the prompt
% tokens under the model's chat template. The decoder's parameters are never updated;
% the only trainable parts are the adapter, the encoder of \S\ref{sec:amortize}, and
% the latents themselves.

where $E$ is the LLM's frozen embedding table and $\mathbf{t}$ the prompt tokens under the model's chat template. The LLM's parameters are never updated. The only trainable parts are the adapter, the encoder of \S\ref{sec:amortize}, and the latents themselves.

\subsection{Latents as free parameters}
\label{sec:glo}

% The usual choice is an autoencoder: a parametric encoder produces
% $\mathbf{z}_i = f_\phi(X_i)$, and the decoder reconstructs the person's text. This
% conflates two failure modes. If the resulting conditioning is weak, one cannot tell
% whether the latent \emph{space} cannot express the person or the \emph{encoder}
% cannot locate them in it. In preliminary experiments the distinction mattered: a
% learned encoder did not improve on a PCA projection of its own inputs at matched
% dimensionality, and no change to the encoder architecture closed the gap.

% We therefore follow generative latent optimization
% \citep{bojanowski2018optimizing}: every training person owns a latent that is a
% \emph{free parameter}, initialized from $\mathcal{N}(0,\sigma_0^2 I)$ and updated by
% gradient descent like any other weight. Training optimizes
% $\{\mathbf{z}_i\}_{i \in \mathcal{P}}$ and $\theta$ jointly. Because no network
% constrains the latent, this family contains every solution an encoder could reach
% and others besides, which removes the encoder as a confound.

%%%%%%% 이 부분은 없어도 됌
The usual choice is an autoencoder, in which a parametric encoder produces
$\mathbf{z}_i = f_\phi(X_i)$ and the LLM reconstructs the person's text.

This setup mixes two possible failure modes. If the resulting conditioning is weak,
it is unclear whether the latent space itself cannot represent the person or whether
the encoder fails to find the right point in that space.

In preliminary experiments this distinction mattered. A learned encoder did not
outperform a PCA projection of the same inputs at matched dimensionality. Changing
the encoder architecture did not remove this gap.
%%%%%%%

We therefore use Generative Latent Optimization \citep{glo}. Each
training person is assigned a latent vector that is treated as a free parameter,
initialized from $\mathcal{N}(0,\sigma_0^2 I)$, where $\sigma_0$ controls the
initialization scale and $I$ is the $d\times d$ identity matrix, and then updated by
gradient descent like any other model parameter. Training jointly optimizes
$\{\mathbf{z}_i\}_{i\in\mathcal{P}}$ and $\theta$.

Because the latent of a training person is optimized directly rather than produced by
a network, it is not restricted by the encoder. For these people the model can reach
any solution available to the encoder and potentially others as well, so the encoder
no longer confounds the question of whether the latent space can represent a person.
For people unseen in training no free latent exists, and we return to the encoder,
whose prediction can then be refined at test time (\S\ref{sec:testtime}).

\paragraph{Reconstruction.}
The primary signal is the likelihood of the person's own posts under the
conditioned LLM, written in the same chat format used at inference, so that
training and inference see the same prefix position:
\begin{equation}
\mathcal{L}_{\text{rec}}
= \mathbb{E}_{x \sim X_i}\big[-\log p(x \mid \mathbf{t}, \mathbf{z}_i)\big].
\end{equation}

\paragraph{Margin.}
Reconstruction alone admits a degenerate solution in which the adapter learns one
generic prefix and ignores $\mathbf{z}$. We require that a person's own latent
explain their text better than another person's latent by a margin $m$:
% \begin{equation}
% \mathcal{L}_{\text{margin}}
% = \big[\, m - \big(\mathrm{NLL}(x \mid \mathbf{z}_{j}) - \mathrm{NLL}(x \mid \mathbf{z}_{i})\big) \,\big]_{+},
% \qquad j \neq i ,
% \end{equation}
\begin{equation}
\begin{aligned}
\mathcal{L}_{\text{margin}}
= \big[\, m &- \mathrm{NLL}(x \mid \mathbf{z}_{j}) \\
&+ \mathrm{NLL}(x \mid \mathbf{z}_{i}) \,\big]_{+}, \quad j \neq i .
\end{aligned}
\end{equation}
where $[\cdot]_+$ is the hinge and $j$ is drawn from the same batch.

\paragraph{Norm control.} Free latents are otherwise unconstrained and drift away from
the region of the space on which the adapter was trained, which lowers likelihood
while degrading generation. We therefore add $\mathcal{L}_{\mathrm{norm}} =
\lVert \mathbf{z}_i \rVert^2$.

\subsection{Amortization: retaining an encoder}
\label{sec:amortize}

Free latents exist only for people seen in training. To generalize we train, in the
same loop, a permutation-invariant set encoder $f_\phi$ that predicts the free
latent from the person's posts. Each post is embedded by a frozen authorship
encoder \citep{luar}, and the resulting set is pooled with learned
queries:
% \begin{equation}
% f_\phi(X_i) = W_o\,\mathrm{Pool}_\phi\big(\{\,\mathrm{Enc}(x_{ik})\,\}_{k=1}^{K}\big),
% \qquad
% \mathcal{L}_{\text{amort}} = \big\lVert f_\phi(X_i) - \mathrm{sg}[\mathbf{z}_i] \big\rVert^2 ,
% \end{equation}
\begin{equation}
\begin{aligned}
f_\phi(X_i) &= W_o\,\mathrm{Pool}_\phi\big(\{\mathrm{Enc}(x_{ik})\}_{k=1}^{K}\big), \\
\mathcal{L}_{\text{amort}} &= \big\lVert f_\phi(X_i) - \mathrm{sg}[\mathbf{z}_i] \big\rVert^2 .
\end{aligned}
\end{equation}

where $\mathrm{sg}[\cdot]$ is a stop-gradient, so amortization chases the latents
without distorting them. Pooling is order-invariant, so the encoder is unaffected by
the order in which a person's posts arrive.

The full training objective is
\begin{equation}
\mathcal{L} = \mathcal{L}_{\text{rec}}
            + \lambda_m \mathcal{L}_{\text{margin}}
            + \lambda_a \mathcal{L}_{\text{amort}}
            + \lambda_n \mathcal{L}_{\text{norm}} ,
\end{equation}
minimized over $\{\mathbf{z}_i\}$, $\theta$ and $\phi$. Two optimizers are used: one
for $\theta,\phi$ and one, with a larger step size, for the free latents.

\subsection{Inference by optimization}
\label{sec:testtime}

For an unseen person the encoder gives a prediction, but nothing forces that
prediction to be the best point in the space. Since the adapter and the LLM are
fixed at test time, we can continue the same objective and optimize only the $d$
coordinates of the latent:
% \begin{equation}
% \mathbf{z}^{\star}_i
% = \operatorname*{arg\,min}_{\mathbf{z}} \;
% \mathbb{E}_{x \sim X_i}\big[-\log p(x \mid \mathbf{t}, \mathbf{z})\big]
% + \lambda_n \lVert \mathbf{z} \rVert^2,
% \qquad
% \mathbf{z}^{(0)} = f_\phi(X_i).
% \end{equation}
\begin{equation}
\begin{aligned}
\mathbf{z}^{\star}_i
&= \operatorname*{arg\,min}_{\mathbf{z}}
-\mathbb{E}_{x \sim X_i}[\log p(x \,|\, \mathbf{t}, \mathbf{z})]
+ \lambda_n \lVert \mathbf{z} \rVert^2, \\
\mathbf{z}^{(0)} &= f_\phi(X_i).
\end{aligned}
\end{equation}
This is a one-time cost per person: $d$ parameters for a few dozen gradient steps,
after which the latent is stored and reused for every subsequent generation. The
encoder prediction remains available when no such cost is acceptable, so the method
offers two operating points without retraining, and the gap between
$f_\phi(X_i)$ and $\mathbf{z}^\star_i$ measures what amortization gives up.

% \begin{algorithm}[t]
% \caption{Training and inference}
% \label{alg:main}
% \begin{algorithmic}[1]
% \item[] \textbf{Training} \COMMENT{decoder frozen throughout}
% \STATE initialize latents $\{\mathbf{z}_i\}$, adapter $\theta$, encoder $\phi$
% \FOR{each step}
%   \STATE sample people $i$; for each, sample posts $X_i$ and target $x$
%   \STATE compute $\mathcal{L}_{\text{rec}}, \mathcal{L}_{\text{margin}}$ with prefix $g_\theta(\mathbf{z}_i)$
%   \STATE compute $\mathcal{L}_{\text{amort}}$ from $f_\phi(X_i)$ and $\mathrm{sg}[\mathbf{z}_i]$
%   \STATE update $\theta, \phi$; update $\{\mathbf{z}_i\}$ with a separate optimizer
% \ENDFOR
% \item[] \textbf{Inference for a new person}
% \STATE $\mathbf{z} \leftarrow f_\phi(X_{\text{new}})$ \COMMENT{one forward pass}
% \STATE optionally refine $\mathbf{z}$ (\S\ref{sec:testtime}) and store it
% \STATE generate with $[\,g_\theta(\mathbf{z});\, E(\mathbf{t})\,]$ for any prompt $\mathbf{t}$
% \end{algorithmic}
% \end{algorithm}

\subsection{LLM-independent latents}
\label{sec:multi}

Nothing in the objective ties $\mathbf{z}$ to one LLM. When several LLMs are
available we keep a single latent per person and give each LLM its own adapter
$g_{\theta^{(m)}}$, selecting one LLM per step:
% \begin{equation}
% \mathcal{L}_{\text{rec}}^{(m)}
% = \mathbb{E}_{x \sim X_i}\big[-\log p^{(m)}(x \mid \mathbf{t}, \mathbf{z}_i)\big],
% \qquad m \sim \mathrm{Unif}\{1,\dots,M\} .
% \end{equation}

\begin{equation}
\mathcal{L}_{\text{rec}}^{(m)}
= \mathbb{E}_{x \sim X_i}\big[-\log p^{(m)}(x \mid \mathbf{t}, \mathbf{z}_i)\big],
\end{equation}
where $m$ is drawn uniformly from $\{1,\dots,M\}$ at each step.
The latent is then a shared coordinate that each adapter translates into its own
model's input space, which lets the same personality vector drive LLMs of
different families and sizes.
% Experimental Setup.

\section{Experimental Setup}
\label{sec:setup}

\subsection{Data}
\label{sec:data}

We train on corpora of person level writing and evaluate against personality that was
documented independently of us. Kaggle MBTI provides $8{,}608$ people with four MBTI
axes taken from self reported forum flair, Essays \citep{essays} provides
$2{,}081$ people with binary Big Five labels from a questionnaire, and PANDORA
\citep{pandora} provides $1{,}178$ people with continuous Big Five scores from
posted test results. People are split 80/10/10 by a hash of the person identifier,
which for PANDORA is the trait tuple described below, so no individual appears in more
than one split, and labels are never used during training. Each person keeps between
$8$ and $64$ posts of at least $80$ characters, and the single long document of each
Essays author is segmented at sentence boundaries into units of at most $320$
characters.

MBTI labels in the Kaggle corpus are encoded as a four bit integer. Decoding each bit
to its axis reproduces the known type distribution of that corpus, with $1{,}819$
INFP, $1{,}462$ INFJ and $1{,}297$ INTP. Training uses PANDORA through a public mirror
that keeps the text and the trait scores but drops all author identifiers, so we group
posts by their trait tuple, which is unique for every person in that corpus, and use
this tuple as the person identifier both for grouping and for the split.

% We train on corpora of person level writing and evaluate against personality that was
% documented. Splits are made
% by person, so no individual appears in more than one split, and labels are never used
% during training.

% % \begin{table}[t]
% % \centering\small
% % \setlength{\tabcolsep}{4pt}
% % \begin{tabular}{lrrl}
% % \toprule
% % Corpus & People & Posts & Labels \\
% % \midrule
% % Kaggle MBTI & 8{,}608  & 313{,}965 & MBTI, 4 axes \\
% % Essays      & 2{,}081  & 27{,}798  & Big Five, binary \\
% % PANDORA     & 1{,}178  & 74{,}079  & Big Five, continuous \\
% % \midrule
% % Total       & 11{,}867 & 415{,}842 & \\
% % \bottomrule
% % \end{tabular}
% % \caption{Corpora. People are split 80/10/10 by a hash of the person identifier. Each
% % person keeps at most 64 posts and needs at least 8, and posts shorter than 80
% % characters are dropped. Essays contains one long document per person, which we segment
% % at sentence boundaries into units of at most 320 characters.}
% % \label{tab:data}
% % \end{table}

% From Big5-Chat dialogues \citep{big5chat} we take the partner's turn as the context and the persona's reply as
% the target, giving $99{,}344$ pairs from $6{,}209$ people. From Essays we take one
% segment as the context and the segment that follows it as the target, giving
% $25{,}717$ pairs from $2{,}081$ people. From RoleBench we take a question as the
% context and the character's answer as the target, giving $6{,}080$ pairs from $95$
% characters.

\subsection{Model and training}

Each post is first turned into a 512 dimensional vector by a frozen LUAR-MUD encoder
\citep{luar}. The set encoder then summarizes up to 32 of these post vectors into a
single prediction of the person's latent. It does so with attention, using four
learned queries that each attend over all posts, followed by a hidden layer of size
512. Each latent has $d=128$ dimensions. Text is generated by a Qwen2.5-3B-Instruct
LLM, which stays frozen throughout. To condition it on a person, the adapter
maps the latent to $P=24$ soft prefix vectors that are placed in front of the prompt,
where they act like a system prompt written in vectors rather than words.

Training runs for $30{,}000$ steps. Each step samples $16$ people and $2$ of their
texts per person as reconstruction targets, truncated to $160$ tokens. The adapter and
encoder are updated with AdamW at a learning rate of $3\times10^{-4}$ under a one cycle
schedule. The free latents are updated separately with Adam at a larger learning rate
of $10^{-2}$, since each latent receives a gradient only when its person is sampled.
The loss weights are $\lambda_m=\lambda_a=1$ and $\lambda_n=10^{-2}$, and the margin is
$m=0.15$.

Training takes about $17$ hours on a single 48\,GB GPU. Because every step draws new
people and new prompts, the training loss fluctuates and does not clearly show
convergence. We therefore also track the loss on a fixed held out batch.

Test time refinement, described in \S\ref{sec:testtime}, starts from the encoder's
prediction for a new person and improves it by gradient descent on that person's own
posts. We run $60$ Adam steps at a learning rate of $5\times10^{-2}$, using four of the
person's posts per step and the same norm penalty as in training. This takes about $30$
seconds per person on one GPU and is done once, after which the refined latent is
stored and reused.

% The experiment in \S\ref{sec:one-latent} tests whether a latent learned with one
% LLM can be used by another. We take the latents and encoder trained with
% Qwen2.5-3B-Instruct, freeze them, and train only a new adapter for
% Llama-3.1-8B-Instruct for $12{,}000$ steps with $12$ people per step, using the
% reconstruction and margin losses. Because the latents receive no gradient, the second
% LLM has no influence on what they contain.

\subsection{Evaluation protocols}

All evaluations use the same LLM and the same held out people, $120$ from each
corpus and $360$ in total. None of these people appear in training.

\paragraph{Behavioral protocols.}
These protocols ask whether a person's personality shows up in what the conditioned
model writes. They proceed in three steps.

First, we build a scorer that reads personality from text. It takes the average
embedding of a person's posts and predicts their label, using scikit-learn logistic regression for
binary labels and ridge regression for continuous ones with default settings. The scorer is trained only on
real posts written by people in the training split, so it learns what personality
looks like in human writing and never sees text produced by any model.

Second, we inject a held out person's latent and ask the model questions. We use two
question sets. The eight neutral prompts describe an everyday situation without
mentioning personality, such as a plan the person was looking forward to being
cancelled at the last minute. The twelve interview questions are designed to reveal a
specific trait through a concrete situation without naming it, such as what the person
does in the first ten minutes at a party where they know only one person. Each answer
is limited to $120$ new tokens and sampled with temperature $0.9$ and nucleus
sampling at $p=0.95$.

Third, we average the embeddings of the model's answers, apply the scorer, and check
across people whether the predicted personality matches the person's documented one.
We repeat all three steps in two further embedding spaces, RoBERTa-large and
all-MiniLM-L6-v2, fitting a new scorer in each.

% \paragraph{Inventory protocol.}
% This protocol asks the model to describe itself on a standard personality
% questionnaire. For the Big Five corpora we use a $30$ item short form of the BFI-2,
% with six items per trait covering its facets. For the MBTI corpus we use a $32$ item
% inventory written for the four axes. Each item is asked as a question to the person,
% for instance whether they see themselves as someone who is outgoing and sociable.

% The model answers in two turns. It first replies in a sentence or two in its own
% words. A second message then asks for a rating from $1$ to $5$. Rather than parsing
% the number the model writes, we read the probabilities it assigns to the tokens
% \texttt{1} through \texttt{5} and take their expected value. This way every item
% receives a score, and none of the $52{,}080$ items fail. 

\paragraph{Latent space analysis.}
This analysis looks at the latents directly, without generating any text. For held out
people, we train a linear probe to predict each trait label from the latent, using
five fold cross validation. The weight vector of each probe gives the direction in
latent space that corresponds to that trait. We also compute the first ten principal
components of the latents and test whether any of them correlates with a trait label,
controlling the false discovery rate~\citep{10.1111/j.2517-6161.1995.tb02031.x} at $0.05$ across all component and label pairs.
These probes are used only for this analysis and not for any protocol above.

% \paragraph{Leakage.}
% A model could match a person's personality simply by imitating them closely. We
% measure how much of the individual, beyond their personality, appears in the generated
% text in two ways.

% The first measure asks whether the author can be identified from style. We use the
% LUAR-MUD encoder, which was trained to tell whether two texts share an author, so
% texts by the same person end up close together in its space. For each person we embed
% the model's answers, find the closest match among the real writing of all $360$
% evaluation people, and report how often that match is the correct person. Random
% guessing would succeed $0.003$ of the time.

% The second measure checks for direct copying. It is the fraction of five word
% sequences in the generated text that also appear in the person's real posts.

\subsection{Baselines}

All baselines use the same frozen LLM, the same people, the same prompts and
the same scoring. The only thing that changes is how the person is represented to the
model. We compare against three kinds of methods.

\noindent\textbf{Persona embeddings.} These methods, like ours, turn a person into a
vector and inject it as a soft prefix. PPlug \citep{liu-etal-2025-llms} combines the
person's posts with weights that depend on the current question, so the vector changes
from question to question. DEP \citep{qiu-etal-2025-latent} represents the person by
how they differ from similar people and compresses this difference with a sparse
autoencoder that keeps only the $k$ strongest features. To make the comparison fair,
both use the same LUAR post embeddings as our method and the same training budget of
$30{,}000$ steps, latent size $d=128$, $24$ prefix tokens, and the same reconstruction
and margin losses. Any difference in results therefore comes from how the person
vector is built. Because the original papers use different tasks and models, we
reimplemented both methods in our setting rather than reproducing their reported
numbers.

\noindent\textbf{Text prompts.} These methods describe the person to the model in
words. TS \citep{10.1145/3701716.3715463} asks the LLM to summarize ten of the
person's posts into a short description, which is then used as the system prompt.
Trait prompting follows Big5-Chat and Machine Mindset \citep{big5chat,cui2023machine}
and states the person's traits in the prompt, using the traits predicted by the same
scorers we use for evaluation. In context places five of the person's own
posts directly in the prompt.

\noindent\textbf{Controls.} Two controls show what happens without real information
about the person. \textsc{shuffle} injects the latent of a different, randomly chosen
person. The model is still conditioned, but on the wrong person, so any score above
this level must come from matching the right person. \textsc{none} uses no
conditioning at all. A method that gives the same answer for everyone scores exactly
$0.5$ on balanced accuracy and AUC.

\subsection{Metrics and statistics}

For binary labels such as MBTI axes we report ROC AUC and balanced accuracy, which
averages the accuracy on each class. We avoid plain accuracy because the labels are
unbalanced. In the MBTI corpus, for example, most people are introverts, so a method
that predicts introvert for everyone would already look accurate. For continuous Big Five scores we report the Pearson correlation $r$ between predicted and documented scores.

All comparisons are paired, meaning every method is evaluated on the same people with
the same prompts and scorers. To measure uncertainty we use a bootstrap over people.
We draw $2{,}000$ resampled sets of people, using the same resampled sets for every
method, and recompute each score on every set. From these we report $95\%$ confidence
intervals and $P(\Delta \le 0)$, the fraction of resamples in which our method does
not outperform the baseline. A small value means our advantage is unlikely to be a
result of which people happened to be sampled.
% Results section. Numbers are from the current runs. Items marked pending
% await the retraining described in \S\ref{sec:context}.
\begin{table}[t]
\centering\small
\setlength{\tabcolsep}{4pt}
\begin{tabular}{llccc}
\toprule
 & & \textbf{MBTI} & \textbf{Essays} & \textbf{PANDORA} \\
\textbf{Type} & \textbf{Method} & \textbf{AUC} & \textbf{AUC} & $\boldsymbol{r}$ \\
\midrule
\rowcolor{gray!15}
\multicolumn{5}{l}{\emph{Neutral generation}} \\
\multirow{2}{*}{PL(Ours)}      & \textbf{Refined $\mathbf{z}$} & \textbf{0.754} & 0.563 & 0.135 \\
                           & Encoder $\mathbf{z}$          & 0.649 & \underline{0.579} & \textbf{0.204} \\
\cmidrule(l){1-5}
\multirow{2}{*}{Embedding} & PPlug                         & \underline{0.661} & 0.561 & \underline{0.203} \\
                           & DEP                           & 0.634 & 0.564 & 0.179 \\
\cmidrule(l){1-5}
\multirow{3}{*}{Prompt}    & Trait prompt                  & 0.651 & \textbf{0.601} & 0.054 \\
                           & TS                            & 0.620 & 0.545 & 0.077 \\
                           & In context                    & 0.592 & 0.560 & 0.084 \\
\cmidrule(l){1-5}
\multirow{2}{*}{Control}   & \textsc{shuffle}              & 0.499 & 0.473 & 0.032 \\
                           & \textsc{none}                 & 0.528 & 0.534 & 0.040 \\
\midrule
\rowcolor{gray!15}
\multicolumn{5}{l}{\emph{Interview}} \\
\multirow{2}{*}{PL(Ours)}      & \textbf{Refined $\mathbf{z}$} & \textbf{0.782} & 0.588 & \underline{0.164} \\
                           & Encoder $\mathbf{z}$          & \underline{0.729} & 0.573 & 0.143 \\
\cmidrule(l){1-5}
\multirow{2}{*}{Embedding} & PPlug                         & 0.632 & \underline{0.589} & \textbf{0.208} \\
                           & DEP                           & 0.643 & \textbf{0.595} & 0.156 \\
\cmidrule(l){1-5}
\multirow{3}{*}{Prompt}    & Trait prompt                  & 0.655 & 0.526 & 0.079 \\
                           & TS                            & 0.578 & 0.532 & 0.118 \\
                           & In context                    & 0.656 & 0.567 & 0.104 \\
\cmidrule(l){1-5}
\multirow{2}{*}{Control}   & \textsc{shuffle}              & 0.468 & 0.513 & 0.034 \\
                           & \textsc{none}                 & 0.556 & 0.487 & $-0.006$ \\
\bottomrule
\end{tabular}
\caption{Personality read from generated text by scorers trained on real human posts
in the LUAR feature space, for 360 held out people. Best is in bold and second best is
underlined. Embedding baselines are PPlug \citep{liu-etal-2025-llms} and DEP
\citep{qiu-etal-2025-latent}. Prompt baselines are trait prompting in the manner of
Big5-Chat \citep{big5chat}, TS \citep{10.1145/3701716.3715463} and five of the
person's posts in context.}
\label{tab:main}
\end{table}

\section{Results}
\label{sec:results}

% We evaluate in two ways. The first checks whether personality shows up in how the
% conditioned model behaves. The model answers neutral prompts and interview questions,
% and we read personality from its answers with scorers trained on real human writing.
% These results appear in \S\ref{sec:res-behaviour}. The second asks the model to fill
% in a personality questionnaire and takes the scores it reports about itself. These
% results appear in \S\ref{sec:res-selfreport}. In every comparison the methods share
% the same LLM agent, people, prompts and scoring, so they differ only in how the person
% is represented.

We test whether personality appears in what the conditioned model writes
(\S\ref{sec:res-behaviour}), whether a latent learned with one LLM can be read by
another (\S\ref{sec:one-latent}), and where personality resides in the latent space
(\S\ref{sec:latent-space}).

\subsection{Personality expressed in behavior}
\label{sec:res-behaviour}

Table~\ref{tab:main} shows how well each method transfers personality into what the
model writes. On the MBTI axes our refined latents score
highest under both kinds of questions, reaching an AUC of $0.754$ on neutral prompts
and $0.782$ on interviews. On Essays and PANDORA the differences between methods are small and no method leads consistently. Our latents nonetheless remain competitive: one of our two
variants ranks among the top three in every column, and on PANDORA the
embedding-based methods, ours included, reach higher correlations than every
prompt-based method under both question sets. The signal is weak overall, however,
with Essays AUCs below $0.61$ and PANDORA correlations below $0.21$.

\begin{figure*}[t]
  \centering
  \includegraphics[width=0.8\textwidth]{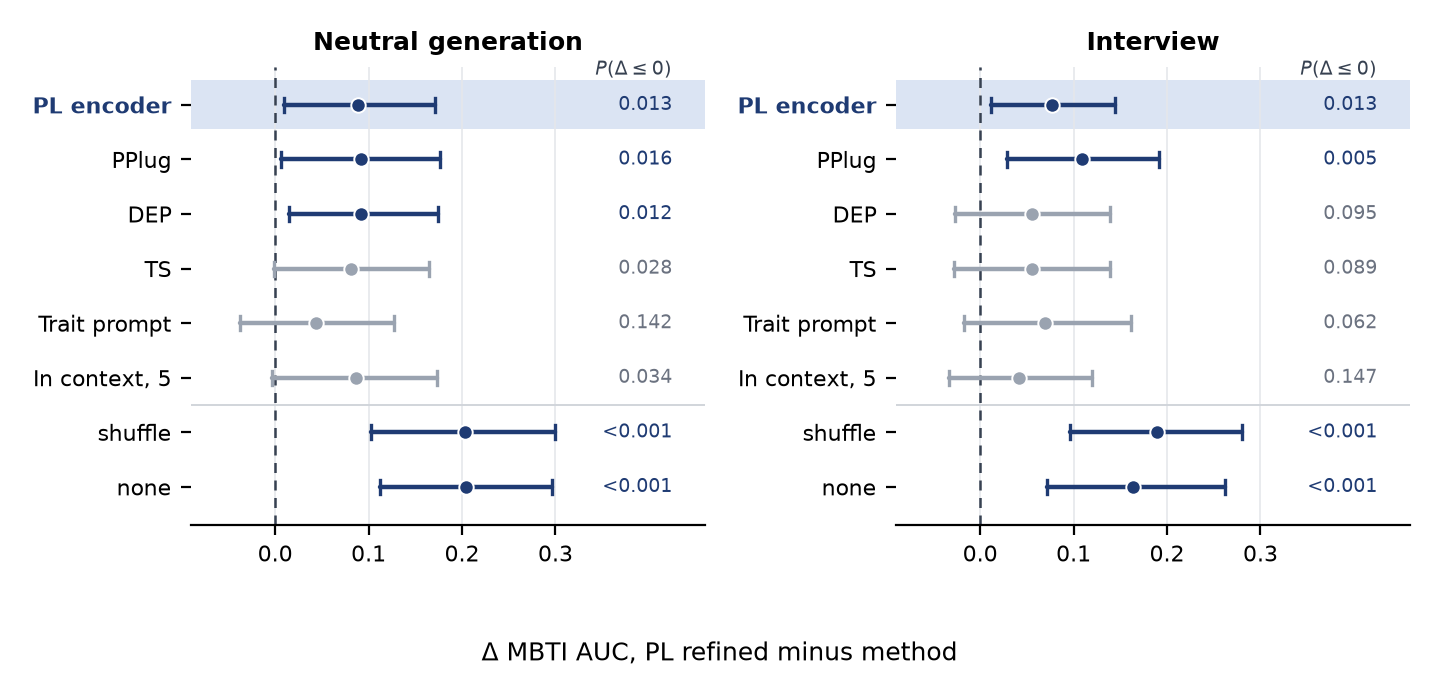}
  \caption{Paired bootstrap over people with 2{,}000 resamples, on MBTI AUC under
RoBERTa-large scoring.}
  \label{fig:forest}
\end{figure*}

\paragraph{The ranking does not depend on the scoring encoder.}
Our latents are built from LUAR, an encoder that captures writing style. If we also
score the answers with LUAR, a method could look good simply by copying the person's
style. To check this, we score the same answers with two encoders that have nothing to
do with authorship, RoBERTa-large and MiniLM, training a new scorer in each, as shown
in Table~\ref{tab:encoders}. The numbers change, and LUAR does make our scores look
somewhat higher, but our method stays first in every space. Figure~\ref{fig:forest} checks
whether these gaps could be due to chance, using the scores from RoBERTa-large, where
our margin is smallest. Our advantage over PPlug is significant on both question sets,
and our advantage over DEP is significant on neutral prompts.

\begin{table}[t]
\centering\small
\setlength{\tabcolsep}{4pt}
\begin{tabular}{llccc}
\toprule
\textbf{Type} & \textbf{Method} & \textbf{LUAR} & \textbf{RoBERTa} & \textbf{MiniLM} \\
\midrule
\rowcolor{gray!15}
\multicolumn{5}{l}{\emph{Neutral generation}} \\
\multirow{2}{*}{PL(Ours)}      & \textbf{Refined $\mathbf{z}$} & \textbf{0.754} & \textbf{0.686} & \textbf{0.792} \\
                           & Encoder $\mathbf{z}$          & 0.649 & 0.597 & 0.638 \\
\cmidrule(l){1-5}
\multirow{2}{*}{Embedding} & PPlug                         & \underline{0.661} & 0.594 & \underline{0.656} \\
                           & DEP                           & 0.634 & 0.595 & 0.587 \\
\cmidrule(l){1-5}
\multirow{3}{*}{Prompt}    & Trait prompt                  & 0.651 & \underline{0.642} & 0.630 \\
                           & TS                            & 0.620 & 0.607 & 0.620 \\
                           & In context                    & 0.592 & 0.601 & 0.623 \\
\cmidrule(l){1-5}
Control                    & \textsc{none}                 & 0.528 & 0.483 & 0.464 \\
\midrule
\rowcolor{gray!15}
\multicolumn{5}{l}{\emph{Interview}} \\
\multirow{2}{*}{PL(Ours)}      & \textbf{Refined $\mathbf{z}$} & \textbf{0.782} & \textbf{0.686} & \textbf{0.790} \\
                           & Encoder $\mathbf{z}$          & \underline{0.729} & 0.611 & \underline{0.724} \\
\cmidrule(l){1-5}
\multirow{2}{*}{Embedding} & PPlug                         & 0.632 & 0.577 & 0.682 \\
                           & DEP                           & 0.643 & 0.632 & 0.650 \\
\cmidrule(l){1-5}
\multirow{3}{*}{Prompt}    & Trait prompt                  & 0.655 & 0.617 & 0.651 \\
                           & TS                            & 0.578 & 0.632 & 0.622 \\
                           & In context                    & 0.656 & \underline{0.645} & 0.655 \\
\cmidrule(l){1-5}
Control                    & \textsc{none}                 & 0.556 & 0.523 & 0.474 \\
\bottomrule
\end{tabular}
\caption{MBTI AUC for the same generations scored by three independent encoders,
LUAR, RoBERTa-large and all-MiniLM-L6-v2. Scorers are refitted on real human posts
within the space of each encoder. Best is in bold and second best is underlined.}
\label{tab:encoders}
\end{table}

% \begin{table}[t]
% \centering\small
% \begin{tabular}{lccc}
% \toprule
% Ours, refined, minus & $\Delta$ & 95\% CI & $P(\Delta \le 0)$ \\
% \midrule
% \multicolumn{4}{l}{\emph{Neutral generation}} \\
% PPlug               & $+0.092$ & $[+0.006, +0.176]$ & $0.016$ \\
% DEP                 & $+0.092$ & $[+0.015, +0.174]$ & $0.012$ \\
% In context, 5 posts & $+0.087$ & $[-0.003, +0.173]$ & $0.034$ \\
% TS                  & $+0.081$ & $[-0.001, +0.165]$ & $0.028$ \\
% Trait prompt        & $+0.044$ & $[-0.038, +0.127]$ & $0.142$ \\
% \textsc{none}       & $+0.204$ & $[+0.112, +0.296]$ & $<0.001$ \\
% \midrule
% \multicolumn{4}{l}{\emph{Interview}} \\
% PPlug               & $+0.109$ & $[+0.029, +0.191]$ & $0.005$ \\
% Ours, encoder       & $+0.077$ & $[+0.012, +0.144]$ & $0.013$ \\
% Trait prompt        & $+0.069$ & $[-0.017, +0.161]$ & $0.062$ \\
% DEP                 & $+0.056$ & $[-0.027, +0.139]$ & $0.095$ \\
% TS                  & $+0.056$ & $[-0.028, +0.139]$ & $0.089$ \\
% In context, 5 posts & $+0.042$ & $[-0.033, +0.120]$ & $0.147$ \\
% \textsc{none}       & $+0.164$ & $[+0.071, +0.262]$ & $<0.001$ \\
% \bottomrule
% \end{tabular}
% \caption{Paired bootstrap over people with 2{,}000 resamples, on MBTI AUC under
% RoBERTa-large scoring.}
% \label{tab:sig}
% \end{table}

\paragraph{Test time refinement adds most of the gain.}
Our method has two modes. The encoder predicts a latent in one step, and test time
refinement then improves it on the person's own posts. With the encoder alone, our
MBTI AUC is $0.649$ on neutral prompts, close to the best baselines, and $0.729$ on
interviews. Refining each latent for sixty steps raises these to $0.754$ and
$0.782$. Under RoBERTa-large scoring this improvement is significant on both question sets,
at $+0.075$ with $P=0.013$. Refinement also changes the latent noticeably, leaving a
cosine similarity of only $0.635$ to the encoder's prediction. This shows that the
encoder does not reach the best latent for each person, and that the remaining gap can
be closed cheaply at test time.

\begin{table}[t]
\centering\small
\begin{tabular}{lccc}
\toprule
\textbf{MBTI AUC} & \textbf{PL(Ours)} & \textsc{shuffle} & \textsc{none} \\
\midrule
\textbf{E/I} & 0.529 & \textbf{0.580} & 0.444 \\
\textbf{S/N}         & \textbf{0.632} & 0.452 & 0.415 \\
\textbf{T/F}          & \textbf{0.642} & 0.509 & 0.412 \\
\textbf{J/P}       & \textbf{0.565} & 0.429 & 0.532 \\
\midrule
\textbf{Mean}                              & \textbf{0.592} & 0.492 & 0.451 \\
\bottomrule
\end{tabular}
\caption{Llama-3.1-8B-Instruct reading latents it never helped to shape, on $120$ held
out MBTI people. The latents and the encoder are those trained with
Qwen2.5-3B-Instruct, and only the adapter is new.}
\label{tab:crossmodel}
\end{table}

\subsection{One latent, Several LLMs}
\label{sec:one-latent}

Our objective does not tie a latent to the LLM it was trained with. To test this, we
keep the latents and the encoder trained with Qwen2.5-3B-Instruct fixed and train only
a new adapter for Llama-3.1-8B-Instruct, for $12{,}000$ steps with $12$ people per step
under the reconstruction and margin losses. No gradient reaches the latents, so Llama
takes no part in shaping them.

Personality still transfers. On held out MBTI people, Llama conditioned on the right
person's latent reaches an AUC of $0.592$. With another person's latent it reaches
$0.492$, and with no conditioning $0.451$. Both gaps are significant. The gain over
\textsc{shuffle} is $+0.100$, with a $95\%$ interval of $[+0.002, +0.198]$ and
$P=0.022$, and the gain over \textsc{none} is $+0.141$, with an interval of
$[+0.048, +0.233]$ and $P=0.004$.

The gain is smaller than with Qwen, which is expected because the latents were shaped
only by Qwen. Still, a model of a different family and size can read the same latent
through nothing more than a small new adapter.

\begin{figure}[t]
  \centering
  \includegraphics[width=1.01\columnwidth]{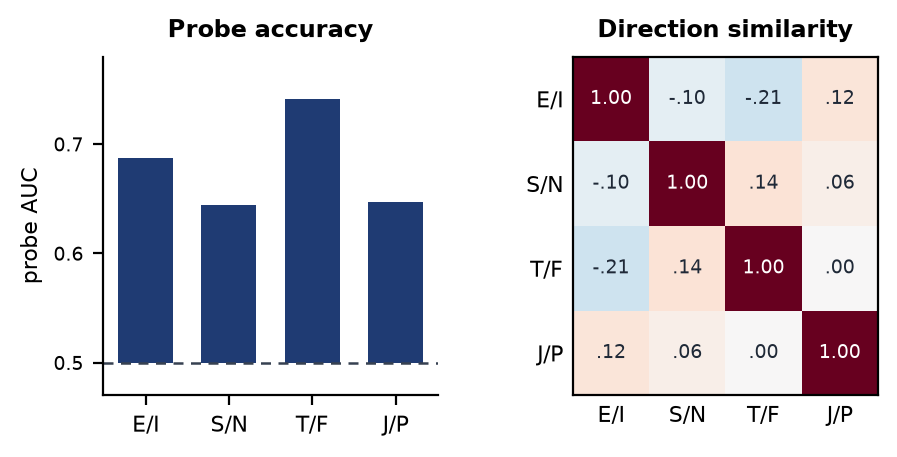}
  \caption{Linear probes on the latents of held out people. Left: probe AUC for each
MBTI axis, with chance at $0.5$. Right: cosine similarity between the four trait
directions, each taken as the weight vector of its probe.}
  \label{fig:auc_cos}
\end{figure}
% \begin{figure}[t]
%   \centering
%   \includegraphics[width=\columnwidth]{fig/fig_cos.png}
%   \caption{Cosine similarity between the four MBTI trait directions, each taken as the
% weight vector of its probe.}
%   \label{fig:cos}
% \end{figure}
% \begin{figure}[t]
%   \centering
%   \includegraphics[width=\columnwidth]{fig/fig_pca.png}
%   \caption{Correlation between the ten leading principal components of the latents and
% each MBTI axis. No pair survives correction at a false discovery rate of $0.05$, and
% the three leading components carry only $22.6\%$ of the variance.}
%   \label{fig:pca}
% \end{figure}

\subsection{What the latent space looks like}
\label{sec:latent-space}

Since no labels were used in training, we can ask afterwards where personality ended
up in the latent space. We first train a simple linear probe to predict each MBTI axis
from the latents of held out people. As the left panel of Figure~\ref{fig:auc_cos}
shows, all four axes can be recovered well above chance, with an AUC of $0.741$ for
thinking versus feeling, $0.687$ for extraversion versus introversion, $0.647$ for
judging versus perceiving and $0.644$ for sensing versus intuition. Each probe also
gives a direction in the space for its trait. The right panel shows that these four
directions are nearly perpendicular to each other, with an average absolute cosine
similarity of $0.105$ and no pair above $0.21$ in magnitude. The traits are therefore
stored largely independently of one another.

% Personality is, however, not the main thing that separates people in the space. We
% used PCA to find the ten directions along which the latents vary most and checked
% whether any of them follows a trait label. Figure~\ref{fig:pca} shows these
% correlations. None of the forty combinations of direction and label is significant
% after correcting for multiple tests with the Benjamini Hochberg procedure at a false
% discovery rate of $0.05$, and the strongest correlation, between the second component
% and the thinking versus feeling axis, is $0.393$. The variance is also spread out,
% with the top three directions explaining only $22.6\%$ of it. In other words, a trait
% is a direction that can be found and used once labels are available, but it is not one
% of the biggest differences between people that the latent encodes.

Personality is, however, not the main thing that separates people in the space. None
of the ten leading principal components of the latents correlates significantly with
any MBTI axis after Benjamini Hochberg correction at a false discovery rate of $0.05$
(Appendix~\ref{app:pca}), so a trait is a direction that labels can find but not one
of the largest differences between people that the latent encodes.

% 내용 더 넣을 수 있으면 넣기
% \subsection{Secondary analyses}

% \paragraph{The latent space has usable axes.}
% Moving a latent along a direction obtained from a probe fitted in latent space shifts
% the corresponding trait in generated text monotonically, at a mean Spearman
% correlation of $0.975$ over the four MBTI axes and an effect size between $0.87$ and
% $0.96$ in predicted probability as $\alpha$ moves from $-2$ to $+2$. The edited axis
% moves $4.8$ times more than the others. The space therefore supports editing and not
% only the retrieval of a stored person.

% \paragraph{Generalization to unseen instruments.}
% On the 24 InCharacter characters with fourteen annotated scales
% \citep{wang2024incharacter}, our latents predict scales never seen in training at a
% mean leave one out correlation of $r=0.156$, against $-0.367$ for a PCA projection at
% matched dimensionality, a difference of $+0.533$ with $P<0.001$. This remains below
% the uncompressed input embedding at $0.348$. Compression therefore retains some
% structure beyond the training taxonomy while discarding part of what the input encoder
% carried.
\section{Conclusion}
\label{sec:conclusion}

We presented \textbf{Personality Latents (PL)}, a way to condition a frozen language model on
a specific individual using only that person's own writing. Each person is represented
by a latent that is optimized directly through the model rather than produced by an
encoder, and an encoder trained alongside makes the method usable for people never
seen in training.

Three findings stand out. First, personality written in a person's own text can be
carried into what a model writes. PL transfers MBTI axes better than persona
embedding and prompt-based baselines, and this holds across three scoring spaces.
Second, refining the latent at test time accounts for most of this advantage, which
shows that an encoder alone does not find the best representation of a person. Third,
the latent is not tied to the model it was trained with, since a different model can
read it through a small new adapter.

The latent space itself suggests where to go next. Each MBTI axis corresponds to a
direction that a linear probe can find, but these directions are not the main source
of variation between people. Making these traits more prominent in the latent, and
training one latent jointly with several models, are natural next steps.
\section*{Limitations}

Our scorers detect signals in generated text that they associate with a trait, which
is not the same as the model holding the person's personality, and they may partly
respond to topic or vocabulary. Repeating the scoring in three embedding spaces
reduces but does not remove this concern. Our clearest gains are on MBTI, whose labels come from self reported forum flair and which has weaker psychometric support than the Big Five, while the signal on the Big Five corpora is weak for every method we tested. All main
results use one frozen model with three billion parameters, and the cross model
experiment covers only one additional model with the latents kept fixed. Test time
refinement also requires the person's own posts and a short optimization run per
person.

The method builds a representation of a specific individual and keeps part of their
style, so it could be misused to impersonate or profile someone without consent. We do
not release latents of real individuals, and any deployment should require the
consent of the person being represented.


\bibliography{custom}

@InProceedings{glo,
  title = 	 {Optimizing the Latent Space of Generative Networks},
  author =       {Bojanowski, Piotr and Joulin, Armand and Lopez-Pas, David and Szlam, Arthur},
  booktitle = 	 {Proceedings of the 35th International Conference on Machine Learning},
  pages = 	 {600--609},
  year = 	 {2018},
  editor = 	 {Dy, Jennifer and Krause, Andreas},
  volume = 	 {80},
  series = 	 {Proceedings of Machine Learning Research},
  month = 	 {10--15 Jul},
  publisher =    {PMLR},
  url = 	 {https://proceedings.mlr.press/v80/bojanowski18a.html}
}

@inproceedings{luar,
    title = "Learning Universal Authorship Representations",
    author = "Rivera-Soto, Rafael A.  and
      Miano, Olivia Elizabeth  and
      Ordonez, Juanita  and
      Chen, Barry Y.  and
      Khan, Aleem  and
      Bishop, Marcus  and
      Andrews, Nicholas",
    editor = "Moens, Marie-Francine  and
      Huang, Xuanjing  and
      Specia, Lucia  and
      Yih, Scott Wen-tau",
    booktitle = "Proceedings of the 2021 Conference on Empirical Methods in Natural Language Processing",
    month = nov,
    year = "2021",
    address = "Online and Punta Cana, Dominican Republic",
    publisher = "Association for Computational Linguistics",
    url = "https://aclanthology.org/2021.emnlp-main.70/",
    doi = "10.18653/v1/2021.emnlp-main.70",
    pages = "913--919"
}

@inproceedings{big5chat,
    title = "{BIG}5-{CHAT}: Shaping {LLM} Personalities Through Training on Human-Grounded Data",
    author = "Li, Wenkai  and
      Liu, Jiarui  and
      Liu, Andy  and
      Zhou, Xuhui  and
      Diab, Mona T.  and
      Sap, Maarten",
    editor = "Che, Wanxiang  and
      Nabende, Joyce  and
      Shutova, Ekaterina  and
      Pilehvar, Mohammad Taher",
    booktitle = "Proceedings of the 63rd Annual Meeting of the Association for Computational Linguistics (Volume 1: Long Papers)",
    month = jul,
    year = "2025",
    address = "Vienna, Austria",
    publisher = "Association for Computational Linguistics",
    url = "https://aclanthology.org/2025.acl-long.999/",
    doi = "10.18653/v1/2025.acl-long.999",
    pages = "20434--20471",
    ISBN = "979-8-89176-251-0"
}

@inproceedings{shao-etal-2023-character,
    title = "Character-{LLM}: A Trainable Agent for Role-Playing",
    author = "Shao, Yunfan  and
      Li, Linyang  and
      Dai, Junqi  and
      Qiu, Xipeng",
    editor = "Bouamor, Houda  and
      Pino, Juan  and
      Bali, Kalika",
    booktitle = "Proceedings of the 2023 Conference on Empirical Methods in Natural Language Processing",
    month = dec,
    year = "2023",
    address = "Singapore",
    publisher = "Association for Computational Linguistics",
    url = "https://aclanthology.org/2023.emnlp-main.814/",
    pages = "13153--13187"
}

@inproceedings{wang-etal-2024-rolellm,
    title = "{R}ole{LLM}: Benchmarking, Eliciting, and Enhancing Role-Playing Abilities of Large Language Models",
    author = "Wang, Noah  and
      Peng, Z.Y.  and
      Que, Haoran  and
      Liu, Jiaheng  and
      Zhou, Wangchunshu  and
      Wu, Yuhan  and
      Guo, Hongcheng  and
      Gan, Ruitong  and
      Ni, Zehao  and
      Yang, Jian  and
      Zhang, Man  and
      Zhang, Zhaoxiang  and
      Ouyang, Wanli  and
      Xu, Ke  and
      Huang, Wenhao  and
      Fu, Jie  and
      Peng, Junran",
    editor = "Ku, Lun-Wei  and
      Martins, Andre  and
      Srikumar, Vivek",
    booktitle = "Findings of the Association for Computational Linguistics: ACL 2024",
    month = aug,
    year = "2024",
    address = "Bangkok, Thailand",
    publisher = "Association for Computational Linguistics",
    url = "https://aclanthology.org/2024.findings-acl.878/",
    doi = "10.18653/v1/2024.findings-acl.878",
    pages = "14743--14777"
}

@article{cui2023machine,
  title={Machine mindset: An mbti exploration of large language models},
  author={Cui, Jiaxi and Lv, Liuzhenghao and Wen, Jing and Wang, Rongsheng and Tang, Jing and Tian, YongHong and Yuan, Li},
  journal={arXiv preprint arXiv:2312.12999},
  year={2023}
}

@inproceedings{bhandari-etal-2026-activation,
    title = "Activation-Space Personality Steering: Hybrid Layer Selection for Stable Trait Control in {LLM}s",
    author = "Bhandari, Pranav  and
      Fay, Nicolas  and
      Selvaganapathy, Sanjeevan  and
      Datta, Amitava  and
      Naseem, Usman  and
      Nasim, Mehwish",
    editor = "Demberg, Vera  and
      Inui, Kentaro  and
      Marquez, Llu{\'i}s",
    booktitle = "Proceedings of the 19th Conference of the {E}uropean Chapter of the {A}ssociation for {C}omputational {L}inguistics (Volume 1: Long Papers)",
    month = mar,
    year = "2026",
    address = "Rabat, Morocco",
    publisher = "Association for Computational Linguistics",
    url = "https://aclanthology.org/2026.eacl-long.300/",
    doi = "10.18653/v1/2026.eacl-long.300",
    pages = "6388--6403",
    ISBN = "979-8-89176-380-7"
}

@inproceedings{10.1145/3701716.3715463,
author = {Ning, Lin and Liu, Luyang and Wu, Jiaxing and Wu, Neo and Berlowitz, Devora and Prakash, Sushant and Green, Bradley and O'Banion, Shawn and Xie, Jun},
title = {User-LLM: Efficient LLM Contextualization with User Embeddings},
year = {2025},
isbn = {9798400713316},
publisher = {Association for Computing Machinery},
address = {New York, NY, USA},
url = {https://doi.org/10.1145/3701716.3715463},
doi = {10.1145/3701716.3715463},
booktitle = {Companion Proceedings of the ACM on Web Conference 2025},
pages = {1219–1223},
numpages = {5},
location = {Sydney NSW, Australia},
series = {WWW '25}
}

@inproceedings{liu-etal-2025-llms,
    title = "{LLM}s + Persona-Plug = Personalized {LLM}s",
    author = "Liu, Jiongnan  and
      Zhu, Yutao  and
      Wang, Shuting  and
      Wei, Xiaochi  and
      Min, Erxue  and
      Lu, Yu  and
      Wang, Shuaiqiang  and
      Yin, Dawei  and
      Dou, Zhicheng",
    editor = "Che, Wanxiang  and
      Nabende, Joyce  and
      Shutova, Ekaterina  and
      Pilehvar, Mohammad Taher",
    booktitle = "Proceedings of the 63rd Annual Meeting of the Association for Computational Linguistics (Volume 1: Long Papers)",
    month = jul,
    year = "2025",
    address = "Vienna, Austria",
    publisher = "Association for Computational Linguistics",
    url = "https://aclanthology.org/2025.acl-long.461/",
    doi = "10.18653/v1/2025.acl-long.461",
    pages = "9373--9385",
    ISBN = "979-8-89176-251-0"
}

@inproceedings{qiu-etal-2025-latent,
    title = "Latent Inter-User Difference Modeling for {LLM} Personalization",
    author = "Qiu, Yilun  and
      Shi, Tianhao  and
      Zhao, Xiaoyan  and
      Zhu, Fengbin  and
      Zhang, Yang  and
      Feng, Fuli",
    editor = "Christodoulopoulos, Christos  and
      Chakraborty, Tanmoy  and
      Rose, Carolyn  and
      Peng, Violet",
    booktitle = "Proceedings of the 2025 Conference on Empirical Methods in Natural Language Processing",
    month = nov,
    year = "2025",
    address = "Suzhou, China",
    publisher = "Association for Computational Linguistics",
    url = "https://aclanthology.org/2025.emnlp-main.536/",
    doi = "10.18653/v1/2025.emnlp-main.536",
    pages = "10599--10617",
    ISBN = "979-8-89176-332-6"
}

@inproceedings{tan-etal-2024-democratizing,
    title = "Democratizing Large Language Models via Personalized Parameter-Efficient Fine-tuning",
    author = "Tan, Zhaoxuan  and
      Zeng, Qingkai  and
      Tian, Yijun  and
      Liu, Zheyuan  and
      Yin, Bing  and
      Jiang, Meng",
    editor = "Al-Onaizan, Yaser  and
      Bansal, Mohit  and
      Chen, Yun-Nung",
    booktitle = "Proceedings of the 2024 Conference on Empirical Methods in Natural Language Processing",
    month = nov,
    year = "2024",
    address = "Miami, Florida, USA",
    publisher = "Association for Computational Linguistics",
    url = "https://aclanthology.org/2024.emnlp-main.372/",
    doi = "10.18653/v1/2024.emnlp-main.372",
    pages = "6476--6491"
}

@inproceedings{tan-etal-2026-instant,
    title = "Instant Personalized Large Language Model Adaptation via Hypernetwork",
    author = "Tan, Zhaoxuan  and
      Zhang, Zixuan  and
      Wen, Haoyang  and
      Li, Zheng  and
      Zhang, Rongzhi  and
      Chen, Pei  and
      Mo, Fengran  and
      Liu, Zheyuan  and
      Zeng, Qingkai  and
      Yin, Qingyu  and
      Jiang, Meng",
    editor = "Liakata, Maria  and
      Moreira, Viviane P.  and
      Zhang, Jiajun  and
      Jurgens, David",
    booktitle = "Proceedings of the 64th Annual Meeting of the {A}ssociation for {C}omputational {L}inguistics (Volume 1: Long Papers)",
    month = jul,
    year = "2026",
    address = "San Diego, California, United States",
    publisher = "Association for Computational Linguistics",
    url = "https://aclanthology.org/2026.acl-long.1081/",
    doi = "10.18653/v1/2026.acl-long.1081",
    pages = "23557--23580",
    ISBN = "979-8-89176-390-6"
}

@article{10.1111/j.2517-6161.1995.tb02031.x,
    author = {Benjamini, Yoav and Hochberg, Yosef},
    title = {Controlling the False Discovery Rate: A Practical and Powerful Approach to Multiple Testing},
    journal = {Journal of the Royal Statistical Society: Series B (Methodological)},
    volume = {57},
    number = {1},
    pages = {289-300},
    year = {1995},
    month = {01},
    issn = {0035-9246},
    doi = {10.1111/j.2517-6161.1995.tb02031.x},
    url = {https://doi.org/10.1111/j.2517-6161.1995.tb02031.x},
    eprint = {https://academic.oup.com/jrsssb/article-pdf/57/1/289/49173396/jrsssb_57_1_289.pdf},
}

@inproceedings{pandora,
    title = "{PANDORA} Talks: Personality and Demographics on {R}eddit",
    author = "Gjurkovi{\'c}, Matej  and
      Karan, Vanja Mladen  and
      Vukojevi{\'c}, Iva  and
      Bo{\v{s}}njak, Mihaela  and
      Snajder, Jan",
    editor = "Ku, Lun-Wei  and
      Li, Cheng-Te",
    booktitle = "Proceedings of the Ninth International Workshop on Natural Language Processing for Social Media",
    month = jun,
    year = "2021",
    address = "Online",
    publisher = "Association for Computational Linguistics",
    url = "https://aclanthology.org/2021.socialnlp-1.12/",
    doi = "10.18653/v1/2021.socialnlp-1.12",
    pages = "138--152"
}

@article{essays,
  title={Linguistic styles: language use as an individual difference.},
  author={James W. Pennebaker and Laura A. King},
  journal={Journal of personality and social psychology},
  year={1999},
  volume={77 6},
  pages={
          1296-312
        },
  url={https://api.semanticscholar.org/CorpusID:29567532}
}

\appendix

\label{sec:appendix}
\section{Principal Components of the Latent Space}
\label{app:pca}

\begin{figure}[h]
  \centering
  \includegraphics[width=\columnwidth]{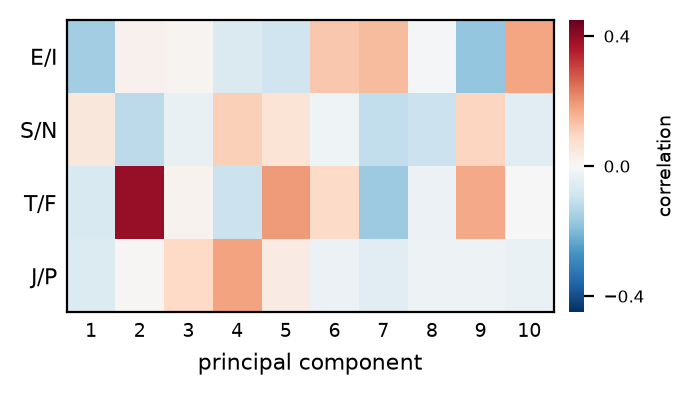}
  \caption{Correlation between the ten leading principal components of the latents and
each MBTI axis. No pair survives correction at a false discovery rate of $0.05$, and
the three leading components carry only $22.6\%$ of the variance.}
  \label{fig:pca}
\end{figure}

We computed the ten directions along which the latents of held out people vary most
and correlated each with the four MBTI axes. Figure~\ref{fig:pca} shows these forty
correlations. None is significant after Benjamini Hochberg correction at a false
discovery rate of $0.05$, and the strongest, between the second component and the
thinking versus feeling axis, is $0.393$. The variance is also spread out, with the top
three components explaining only $22.6\%$ of it.

% This is an appendix.

\end{document}